\documentclass[letterpaper, 10 pt, conference]{ieeeconf}

\IEEEoverridecommandlockouts
\usepackage{microtype}
\usepackage{graphicx}
\usepackage{subfigure}
\usepackage{multirow}
\usepackage{makecell}
\usepackage{booktabs}
\usepackage{siunitx}
\usepackage{wrapfig}
\usepackage{amsmath}
\usepackage{amssymb}
\usepackage{mathtools}
\usepackage{tikz}
\usepackage{tikz-cd}
\usepackage{cite}
\usepackage{placeins}
\usepackage{capt-of}
\usepackage{xcolor}

\title{\LARGE \bf
Acting from Belief, Looking When Needed: A Bayesian Spatial World Model for Navigation under Intermittent Perception\vspace{-0.45em}
}

\author{Feihong Yang$^{1,\dagger}$, Xiang Long$^{1,\dagger}$, Jincheng Yu$^{1,2}$, Jianfei Zhang$^{1}$,
Guangjun Ge$^{1}$, Chao Wang$^{1,*}$, and Yu Wang$^{1,2}$%
\thanks{$^{1}$All authors are with Qiyuan Lab, Beijing, China. $^{2}$Jincheng Yu and Yu Wang are also with the Department of Electronic Engineering, Tsinghua University, Beijing, China. $^{\dagger}$Feihong Yang and Xiang Long contributed equally to this work. $^{*}$Corresponding author: Chao Wang (wangchao1@qiyuanlab.com).}%
}

\IEEEaftertitletext{%
\vspace{-1.35em}%
\begin{minipage}{\textwidth}
\centering
\includegraphics[width=\textwidth]{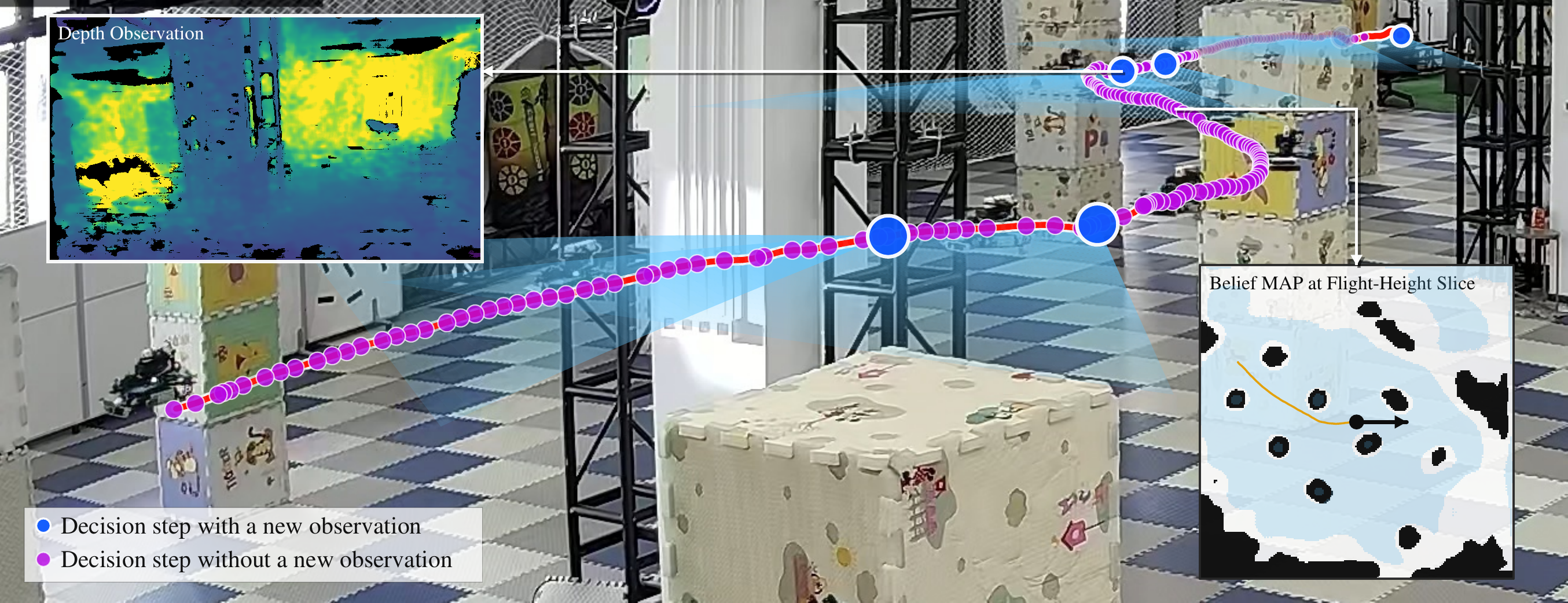}
\captionof{figure}{\textbf{ALONE for drone navigation.}
Indoor real-world flight experiment among regularly arranged obstacles: ALONE continues selecting and executing actions from a maintained spatial belief, requesting depth only when reliable next-action selection needs new evidence. The red curve is the flight path; blue/purple markers denote decision steps with/without a new depth observation, respectively; blue sectors show the corresponding camera views. Insets show an acquired depth image (left) and the spatial estimate decoded from the maintained belief (right).}
\label{fig:exp_overview}
\end{minipage}
\vspace{0.2em}%
}

\begin{document}

\bstctlcite{IEEEexample:BSTcontrol}

\maketitle
\thispagestyle{empty}
\pagestyle{empty}

\begin{abstract}
Robot navigation commonly uses wide-coverage, high-frequency sensing to reduce partial observability; this reliance becomes restrictive when another task temporarily redirects a shared sensor from navigation, interrupting navigation-relevant observations. We study navigation under intermittent perception: acting from an internal spatial belief and looking again only when execution needs a new observation, potentially freeing the shared sensor for other tasks between navigation observations. ALONE, a Bayesian spatial world model, propagates a structured spatial belief using executed actions and corrects it with selectively acquired observations; learned priors over common geometric structures infer unobserved structure from available observation history. It decodes the belief into a spatial estimate for the motion-planning module and predicts a reliability map expressing confidence in the estimate's accuracy. ALONE requests an observation only if insufficient reliability hinders navigation and new evidence should make relevant-region spatial information more reliable; otherwise, it continues acting from the propagated belief. We instantiate ALONE for drone navigation with intermittent single-camera depth images. Across two simulated scene families, it achieves 98\% and 97\% closed-loop success at a 10 Hz decision rate. Among successful trials, median fractions of decision steps requiring a new depth observation are only 0.9\% and 1.3\%, respectively, demonstrating high navigation success with substantially reduced observation demand. Real-world indoor flight experiments further validate navigation under intermittent depth observations, with all 10 trials successful.
\end{abstract}

\section{Introduction}
\label{introduction}
Most robot navigation systems treat perception as an always-on service. To reduce partial observability, they enlarge sensing coverage, observe at high frequency, fuse multiple sensors, and continually update a local map or state estimate~\cite{elfes1989occupancy,hornung2013octomap,mirchev2021variational,mirchev2023prism}. This perception-first strategy is effective, but assumes that sensing can remain continuously directed toward navigation. The assumption becomes restrictive when another task redirects a shared sensor, interrupting navigation-relevant observations.

Humans provide a familiar counterexample. A person can continue walking while visual attention briefly shifts to another task, relying on recent observations, self-motion, and priors to maintain an internal understanding of the surroundings. Such natural gaze behavior is task-dependent and is directed toward the information needed for ongoing action~\cite{land2006eye,hayhoe2005eye}. How long this understanding remains useful depends on environmental predictability. A corridor supports predictable continuation and thus longer observation-free motion, whereas irregular forest geometry leaves upcoming obstacles weakly constrained. These priors support inference between observations; perception returns when reliability is insufficient for the next action.

This observation motivates navigation under intermittent perception: can a robot act from an internal spatial belief and look again only when execution requires new evidence? To this end, the robot must infer current spatial structure from partial action--observation history, preserve alternatives not resolved by available evidence, and determine where its estimate is accurate enough for the motion-planning module's current query. The observation policy should request new evidence only when insufficient reliability hinders navigation and an observation is expected to improve the relevant region.

Existing methods provide individual pieces of this capability. Mapping and spatial memory preserve acquired evidence~\cite{elfes1989occupancy,hornung2013octomap,mirchev2021variational,mirchev2023prism,kayalibay2022tracking,parisotto2017neuralmap,chaplot2020neural,liu2024volumetric,wang2023esceme,chen2025mapnav}; learned world models propagate latent states and predict hidden or future structure~\cite{lecun2022path,hafner2019planet,hafner2025dreamerv3,bar2025nwm,assran2025vjepa2,zhou2025persistent,zheng2024occworld}; and belief-space planning or active perception adapts motion and sensing to uncertainty~\cite{platt2010belief,atanasov2014information,vandenberg2011lqgmp,bajcsy1988active}. They do not by themselves provide one learned spatial belief that is propagated between observations, decoded into a spatial estimate and reliability map, and corrected by requested evidence. Prediction, action, and observation therefore remain separate rather than forming a closed loop.

We call the resulting framework \textbf{ALONE}, reflecting its two operating modes: \textbf{A}cting from belief while perception is suspended and \textbf{L}ooking \textbf{O}nly when \textbf{N}eeded for \textbf{E}xecution. At its core, a decision-queryable Bayesian spatial world model maintains a structured spatial belief. Its action-conditioned prior performs learned spatial completion: learned geometric priors infer unobserved structure from the partial action--observation history. An observation-conditioned posterior corrects this completion when evidence returns. The belief yields a spatial estimate for the motion-planning module and a reliability map indicating confidence in its accuracy at each queried location. We term this probability query-space reliability. ALONE requests an observation only when insufficient reliability restricts navigation and the relevant region is expected to be visible; otherwise, it acts from the maintained belief.

We instantiate ALONE for drone navigation using intermittent depth images from a single camera and a vector-quantized occupancy representation. ALONE preserves high navigation success while reducing observation demand by up to two orders of magnitude. Open-loop and ablation experiments further validate spatial-belief prediction and correction, learned spatial completion, query-space reliability, and the observation policy.

\emph{Contributions.}
Our contributions are threefold: (1) a Bayesian spatial world model that maintains a structured spatial belief through action-conditioned propagation and observation-conditioned correction, using learned geometric priors to infer unobserved structure and providing a spatial estimate together with an explicitly supervised reliability map for planning; (2) a navigation policy that uses spatial reliability to select motion actions and determine when new observations are needed, allowing the robot to continue acting from its belief between observations; and (3) an implementation for drone navigation using a single depth camera, evaluated through open-loop spatial inference, closed-loop navigation comparisons, and real-world flight experiments. The results demonstrate high navigation success with substantially reduced observation demand.

\section{Related Work}
\label{related_work}

\subsection{Spatial Mapping and Memory}
Occupancy mapping integrates measurements into persistent free-space and occupancy estimates~\cite{elfes1989occupancy,hornung2013octomap}, while probabilistic spatial models jointly estimate maps and agent states~\cite{mirchev2021variational,mirchev2023prism,kayalibay2022tracking}. Neural, topological, volumetric, episodic, and semantic memories encode task-relevant spatial structure~\cite{parisotto2017neuralmap,chaplot2020neural,liu2024volumetric,wang2023esceme,chen2025mapnav}. These approaches primarily organize acquired evidence rather than propagate unresolved spatial hypotheses through observation-free motion or assess whether the resulting local structure supports a motion-planning query. ALONE instead propagates a learned categorical spatial belief with executed actions and corrects it with later observations.

\subsection{Predictive World Models}
Latent world models learn action-conditioned dynamics for planning and behavior learning~\cite{hafner2019planet,hafner2025dreamerv3}. Recent work extends prediction to navigation, visual representations, persistent 3D scenes, and occupancy forecasting~\cite{bar2025nwm,assran2025vjepa2,zhou2025persistent,zheng2024occworld}. Yet prediction alone does not establish whether a spatial estimate reliably supports current action selection. ALONE uses learned geometric priors to infer unobserved structure and predicts a reliability map for the estimate, so the same belief supports both action selection and observation decisions.

\subsection{Uncertainty-Aware Planning and Active Perception}
Belief-space planning incorporates state uncertainty into trajectory optimization~\cite{platt2010belief,vandenberg2011lqgmp}, while active perception selects sensing actions or viewpoints using information gain or task-relevant uncertainty~\cite{atanasov2014information,bajcsy1988active}. These methods commonly assume that a suitable belief or uncertainty estimate is available. ALONE instead learns, propagates, queries, and corrects the spatial belief underlying its observation decision. It requests an observation when insufficient reliability restricts navigation and the responsible region is expected to be visible; the returned evidence corrects the maintained belief, closing the loop among prediction, action, observation, and correction.

\section{ALONE: A Bayesian Spatial World Model for Intermittent Perception}
\label{proposed_approach}
\label{sec:framework}

\setlength{\floatsep}{0.55\baselineskip plus 0.1\baselineskip minus 0.1\baselineskip}
\setlength{\textfloatsep}{0.85\baselineskip plus 0.1\baselineskip minus 0.15\baselineskip}
\setlength{\intextsep}{0.70\baselineskip plus 0.1\baselineskip minus 0.1\baselineskip}
\setlength{\dblfloatsep}{0.55\baselineskip plus 0.1\baselineskip minus 0.1\baselineskip}
\setlength{\dbltextfloatsep}{0.85\baselineskip plus 0.1\baselineskip minus 0.15\baselineskip}
\setlength{\abovecaptionskip}{0.30\baselineskip}

\begin{figure*}[!t]
\centering
\includegraphics[width=\textwidth]{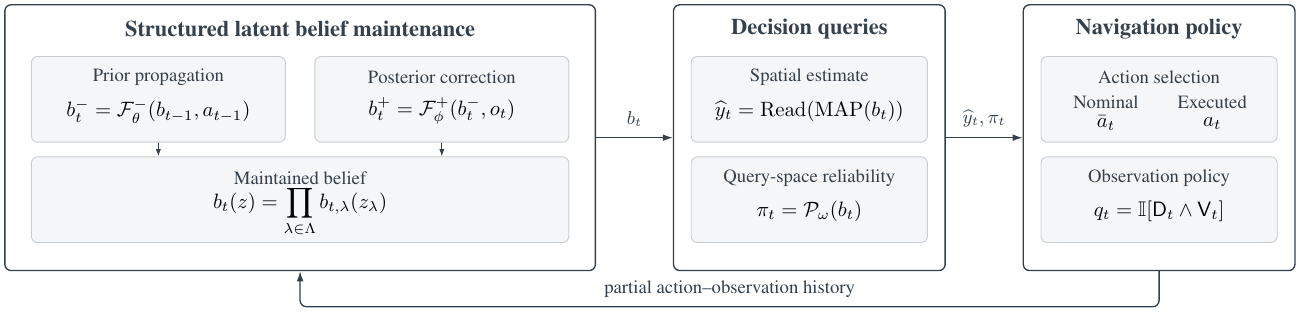}
\caption{\textbf{ALONE framework.} Executed actions propagate a structured spatial belief, and selectively acquired observations correct it. A decoded spatial estimate is provided to the motion-planning module, while the navigation policy uses a reliability map for action selection and the observation decision.}
\label{fig:alone_framework}
\end{figure*}

ALONE has two coupled modes. Without a navigation observation, it propagates the belief with the executed action; the motion-planning module generates candidates from the decoded estimate, and the navigation policy selects one to execute. When insufficient reliability restricts navigation and new evidence is expected to resolve the relevant region, ALONE resumes perception and corrects the belief before the next decision. Observation intervals thus adapt to environmental predictability and decision-relevant uncertainty. We next formulate this setting, describe belief prediction and correction, and define the navigation policy.

\subsection{Navigation under Intermittent Perception}
\label{sec:problem_formulation}

We consider navigation in a partially observable environment where another task may redirect a shared sensor. At each decision step, the navigation policy selects an action to execute and decides whether to acquire a navigation observation. Let \(X_t\in\mathcal X\) denote the navigation-relevant spatial state, \(a_t\in\mathcal A\) the executed action, \(O_t\) an available observation, and \(q_t\in\{0,1\}\) whether that observation is acquired. With \(\mathcal T_t=\{s<t:q_s=1\}\), the information available at decision step \(t\) and its spatial belief are
\begin{equation}
\mathcal I_t=(a_{0:t-1},o_{\mathcal T_t}),\qquad
p_t^X=\mathbb P\!\left(X_t\in\cdot\mid\mathcal I_t\right)
\in\Delta(\mathcal X).
\label{eq:conditional_spatial_belief}
\end{equation}
A navigation policy \((\mu,\nu)\) jointly produces \(a_t=\mu(\mathcal I_t,g)\) and \(q_t=\nu(\mathcal I_t,g)\) for goal \(g\). Let \(J_{\mathrm{nav}}\) denote expected cumulative motion and terminal task cost, and let \(J_{\mathrm{obs}}\) denote the expected total number of acquired observations. We seek
\begin{equation}
\min_{\mu,\nu}\ J_{\mathrm{nav}}(\mu,\nu)
+\lambda_{\mathrm{obs}}J_{\mathrm{obs}}(\nu),
\qquad \lambda_{\mathrm{obs}}>0.
\label{eq:joint_navigation_observation_objective}
\end{equation}
Here \(\lambda_{\mathrm{obs}}>0\) represents the opportunity cost of assigning the shared sensor to navigation. Directly solving this joint belief-space problem is generally intractable. ALONE instead maintains a tractable spatial belief, acts where its estimate is reliable, and requests an observation when insufficient reliability changes the preferred action and new evidence is expected to improve the relevant region.

\subsection{Acting from Belief: Bayesian Spatial Prediction and Correction}
\label{sec:lone_method}

Let \(Y_t=\psi(X_t)\in\mathcal Y\) be the spatial quantity queried by the motion-planning module. ALONE represents \(X_t\) through structured categorical variables \(Z_t=\mathrm{Struct}(X_t)=(Z_{t,\lambda})_{\lambda\in\Lambda}\), where \(Z_{t,\lambda}\in\mathcal C_\lambda\) summarizes a multicoordinate spatial pattern. A decoder \(\mathrm{Read}\) maps a latent hypothesis to the query space, with \(\mathrm{Read}(\mathrm{Struct}(X_t))\approx\psi(X_t)\). For \(\sigma\in\{-,+\}\), corresponding to the pre- and post-observation histories \(\mathcal I_t^-\) and \(\mathcal I_t^+\), respectively, ALONE maintains categorical marginals and the associated factorized approximation
\begin{equation}
 b_{t,\lambda}^\sigma(k)\approx
 \mathbb P\!\left(Z_{t,\lambda}=k\mid\mathcal I_t^\sigma\right),
 \qquad
 b_t^\sigma(z)=\prod_{\lambda\in\Lambda}b_{t,\lambda}^\sigma(z_\lambda).
\label{eq:factorized_latent_belief}
\end{equation}
Each marginal retains alternatives over the spatial pattern represented by one latent variable; the product approximation discards only dependencies across latent variables. This reduces the representation from an intractable joint field distribution to \(O(\sum_\lambda|\mathcal C_\lambda|)\) probabilities.

Executed actions propagate a prior belief, and an acquired observation corrects it:
\begin{equation}
 b_t^- = \mathcal F_\theta^-(b_{t-1},a_{t-1}),
 \qquad
 b_t^+ = \mathcal F_\phi^+(b_t^-,o_t).
\label{eq:learned_filter_maps}
\end{equation}
The maintained belief is \(b_t=b_t^+\) when \(o_t\) is available and \(b_t=b_t^-\) otherwise.
Given supervisory observation \(o_t^{\mathrm{sup}}\), let \(z_t^{\mathrm{sup}}=\mathrm{Sup}_{Z}(o_t^{\mathrm{sup}})\). For \(\sigma\in\{-,+\}\), let \(n_t^\sigma\) denote the number of observation-free inference steps associated with candidate belief \(b_t^\sigma\), with \(n_t^+=0\). Both candidates are supervised through
\begin{equation}
 \mathcal L_{\mathrm{str}}
 =\sum_{t=1}^{T}\sum_{\sigma\in\{-,+\}}
 \alpha_{n_t^\sigma}
 \operatorname{CE}(b_t^\sigma,z_t^{\mathrm{sup}}).
\label{eq:structural_objective}
\end{equation}
Training uses deployment-matched recurrent histories containing both observed and observation-free steps; \(\alpha_n\) balances the contribution of beliefs with \(n\) observation-free inference steps. Categorical log loss is strictly proper, so its population optimum recovers the corresponding conditional categorical marginals when \(\mathrm{Sup}_{Z}\) is consistent with the maintained structural abstraction. Thus \(\mathcal F_\theta^-\) learns prediction from action--observation history, while \(\mathcal F_\phi^+\) learns correction from newly acquired evidence.

\subsection{Looking When Needed: Observation Policy}
\label{sec:navigation_policy}

ALONE derives two complementary outputs from the maintained belief: a spatial estimate used by the motion-planning module to generate candidate actions, and a reliability map used by the navigation policy to determine whether those actions are sufficiently supported and whether new evidence is needed.

\paragraph{Spatial estimate}
The maintained belief is converted into a spatial estimate by componentwise MAP decoding:
\begin{equation}
 z_{t,\lambda}^*=\arg\max_{k\in\mathcal C_\lambda}b_{t,\lambda}(k),
 \qquad
\widehat y_t=\mathrm{Read}(z_t^*).
\label{eq:spatial_estimate}
\end{equation}
This spatial estimate provides the representation on which the motion-planning module generates and evaluates candidate actions.

\paragraph{Query-space reliability}
MAP decoding selects a single structured hypothesis and therefore does not preserve the probabilities assigned to alternatives. ALONE retains the full categorical belief to predict where the decoded estimate satisfies the motion-planning module's accuracy requirement. Let fields in \(\mathcal Y\) be indexed by \(\Omega_{\mathcal Y}\), with pointwise discrepancy \(d_{\mathcal Y}\) and admissible error threshold \(\delta_{\mathcal Y}\). Given supervisory observation \(o_t^{\mathrm{sup}}\), let \(y_t^{\mathrm{sup}}=\mathrm{Sup}_{Y}(o_t^{\mathrm{sup}})\). For \(\sigma\in\{-,+\}\),
\begin{equation}
\begin{aligned}
\widehat y_t^\sigma&=\mathrm{Read}\!\left(\operatorname{MAP}(b_t^\sigma)\right),\\
c_t^{\mathrm{sup},\sigma}(\xi)&=\mathbb I\!\left[
d_{\mathcal Y}\!\left(\widehat y_t^\sigma,y_t^{\mathrm{sup}}\right)(\xi)
\leq\delta_{\mathcal Y}(\xi)\right].
\end{aligned}
\label{eq:supervised_query_adequacy}
\end{equation}
The binary target \(c_t^{\mathrm{sup},\sigma}(\xi)\) labels whether the estimate decoded from the prior or posterior belief is accurate enough at query location \(\xi\).
A decoder \(\mathcal P_\omega\) predicts the adequacy probability \(\pi_t^\sigma=\mathcal P_\omega(b_t^\sigma)\in[0,1]^{\Omega_{\mathcal Y}}\) and is trained with
\begin{equation}
 \mathcal L_{\mathrm{rel}}
 =\sum_{t=1}^{T}\sum_{\sigma\in\{-,+\}}
 \gamma_{n_t^\sigma}
 \operatorname{BCE}\!\left(\pi_t^\sigma,c_t^{\mathrm{sup},\sigma}\right).
\label{eq:adequacy_objective}
\end{equation}
Here \(\gamma_n\) balances reliability supervision across beliefs with \(n\) observation-free inference steps.
Because binary cross-entropy is strictly proper, \(\pi_t(\xi)\) estimates the conditional probability that the spatial estimate meets the accuracy criterion at query location \(\xi\), provided the supervisory query target is consistent. We use this probability directly as query-space reliability. Unlike latent entropy, it indicates whether the spatial estimate is accurate enough for the motion-planning module's current query. Geometrically regular regions can therefore remain admissible across longer observation gaps, whereas weakly constrained regions lose reliability sooner.

\paragraph{Reliability-gated action and observation}
At each decision step, given \(\widehat y_t\) and goal \(g\), the motion-planning module returns a nonempty feasible action set \(\mathcal A_t^0\) containing a designated fallback \(a^{\mathrm{stop}}\), together with a task cost \(\kappa_t\) and, for each candidate \(a\), a nonempty query region \(\Omega(a)\subseteq\Omega_{\mathcal Y}\) swept by the robot during execution. ALONE's navigation policy selects the executed action from this module output using reliability. The nominal action, reliability-admissible set, and executed action are
\begin{equation}
\begin{aligned}
\bar a_t&\in\arg\min_{a\in\mathcal A_t^0}\kappa_t(a),\\
\mathcal A_t^\tau&=\left\{a\in\mathcal A_t^0:
\min_{\xi\in\Omega(a)}\pi_t(\xi)\geq\tau\right\},\\
a_t&=\begin{cases}
\displaystyle\arg\min_{a\in\mathcal A_t^\tau}\kappa_t(a),
&\mathcal A_t^\tau\neq\varnothing,\\
a^{\mathrm{stop}},&\mathcal A_t^\tau=\varnothing.
\end{cases}
\end{aligned}
\label{eq:nominal_supported_sets}
\end{equation}
Thus, \(a_t\) is the lowest-cost candidate whose entire query region meets the reliability threshold, or \(a^{\mathrm{stop}}\) if no candidate does. The selected action is executed and used to propagate the next prior.

The navigation policy sets the binary observation decision \(q_t\) to one only when both of the following conditions hold: \(\mathsf D_t\) indicates that the difference between \(\bar a_t\) and \(a_t\) materially degrades the nominal action, and \(\mathsf V_t\) indicates that the responsible low-reliability region is expected to be observable:
\begin{equation}
q_t=\mathbb I\!\left[
\mathsf D_t(\bar a_t,a_t)=1
\ \land\ 
\mathsf V_t(\Omega_t^{\mathrm{low}},\Omega_t^{\mathrm{vis}})=1
\right].
\label{eq:observation_policy}
\end{equation}
When \(q_t=0\), no navigation observation is acquired. When \(q_t=1\), the requested observation corrects the belief before the next decision step. This closes the loop among belief propagation, action selection, selective observation, and belief correction.

\section{ALONE for Depth-Based Drone Navigation}
\label{sec:drone_implementation}

We instantiate ALONE for collision-aware drone navigation using intermittent depth images from a single camera. At each step, the motion-planning module generates candidate actions from the decoded spatial estimate, and ALONE's navigation policy uses reliability to select the executed action and decide whether to request depth. Executed motion propagates the belief, while requested depth corrects it.

\subsection{Occupancy Representation and Supervision}
\label{sec:drone_state}

The navigation state and motion-planning query are binary occupancy fields in an ego-centric, yaw-aligned voxel frame with a gravity-aligned vertical axis: \(\mathcal X=\mathcal Y=\{0,1\}^{n_x\times n_y\times n_z}\) and \(\psi=\mathrm{Id}\). Actions are relative 3-D poses and runtime observations are depth images. Simulation supplies dense aligned occupancy as \(O^{\mathrm{sup}}\in\mathcal Y\), so \(\mathrm{Sup}_{Y}=\mathrm{Id}\); zero--one discrepancy with a zero error threshold makes adequacy equivalent to voxelwise occupancy agreement. With real data, \(\mathrm{Sup}_{Y}\) may instead construct a query-space target from lidar point clouds or another supervisory sensor.

Following OccWorld~\cite{zheng2024occworld}, a standard vector-quantized autoencoder maps a dense occupancy field to an \(h\times w\) code map with codebook size \(K\) and decodes it back to occupancy. Each latent location summarizes a multivoxel receptive field, allowing one categorical marginal to represent alternative local shapes. Accordingly, \(\mathrm{Struct}=\mathrm{Sup}_{Z}=\mathrm{Enc}_{\mathrm{vq}}\) and \(\mathrm{Read}=\mathrm{Dec}_{\mathrm{vq}}\). The maintained belief is a \(K\times h\times w\) tensor whose componentwise MAP code map is decoded as the spatial estimate \(\widehat y_t\). The adequacy decoder provides its reliability map \(\pi_t=\mathcal P_\omega(b_t)\).

\subsection{Belief Prediction, Correction, and Training}
\label{sec:drone_belief_filter}

Between observations, \(\mathcal W_{a_{t-1}}\) transports the categorical belief into the current ego-centric frame. The prior network completes unobserved structure and predicts categorical logits. When depth becomes available, an observation encoder extracts new evidence and a posterior network fuses it with the prior:
\begin{equation}
\begin{aligned}
b_t^-&=\operatorname{softmax}_{K}\Psi_\theta^-\!\left(
\mathcal W_{a_{t-1}}b_{t-1}\right),\\
b_t^+&=\operatorname{softmax}_{K}\Psi_\phi^+\!\left(
b_t^-,\Phi_{\mathrm{obs}}(o_t)\right).
\end{aligned}
\label{eq:tensor_posterior_impl}
\end{equation}
The softmax is applied over the \(K\) codes at every latent location. The geometric transport \(\mathcal W_{a_{t-1}}\) represents the categorical belief by sampled latent hypotheses. Each hypothesis is decoded to 3-D occupancy, rigidly transformed into the current ego-centric frame, and re-encoded and quantized; the resulting codes form an empirical categorical belief. The observation encoder projects depth through the camera and noise models into probabilistic surface evidence and maps it to latent logits for posterior fusion. The maintained belief is \(b_t^+\) when depth is incorporated and \(b_t^-\) otherwise.

Training proceeds in three stages. First, the VQ autoencoder is trained and frozen, fixing \(\mathrm{Struct}\), \(\mathrm{Sup}_{Z}\), and \(\mathrm{Read}\). Second, \(\Psi_\theta^-\) and \(\Psi_\phi^+\) are trained recurrently with Eq.~\eqref{eq:structural_objective}. A sampled mask generates both observed and observation-free histories by selecting the belief passed to the next step, while both candidates remain supervised:
\begin{equation}
M_t\sim\operatorname{Bernoulli}(p_{\mathrm{obs}}),\qquad
b_t=\begin{cases}
b_t^+,&M_t=1,\\
b_t^-,&M_t=0.
\end{cases}
\label{eq:masked_filter_update}
\end{equation}
Third, the filter is frozen and \(\mathcal P_\omega\) is trained with Eq.~\eqref{eq:adequacy_objective}, using labels generated separately from prior and posterior spatial estimates. The coefficients \(\alpha_{n_t^\sigma}\) and \(\gamma_{n_t^\sigma}\) first average losses among beliefs with the same number of observation-free inference steps and then across these groups, preventing shorter horizons from dominating training.

\subsection{Reliability-Gated Navigation Policy}
\label{sec:drone_planning}

The motion-planning module computes a reference path on \(\widehat y_t\) using A*. Local motion primitives provide candidate actions; each primitive's swept voxels instantiate \(\Omega(a)\), and collision-free primitives form \(\mathcal A_t^0\). A composite cost combining reference-path following and obstacle clearance defines \(\kappa_t\). ALONE's navigation policy then selects the executed action according to Eq.~\eqref{eq:nominal_supported_sets}.

Let \(\Omega_t^{\mathrm{vis}}\) denote the region where a depth observation is expected to provide new evidence from the current pose. It contains voxels inside the depth frustum reached before any intervening voxel \(\zeta\) satisfying \(\widehat y_t(\zeta)=1\) and \(\pi_t(\zeta)\geq\tau\); low-reliability occupancy is not treated as confirmed occlusion. The unresolved region relevant to the observation decision is the low-reliability portion of the nominal action's query support:
\begin{equation}
\Omega_t^{\mathrm{low}}
=\{\xi\in\Omega(\bar a_t):\pi_t(\xi)<\tau\}.
\label{eq:drone_query_quantities}
\end{equation}
The abstract action-impact and observability conditions in Eq.~\eqref{eq:observation_policy} are instantiated as
\begin{equation}
\begin{aligned}
\mathsf D_t
&=\mathbb I\!\left[
\kappa_t(a_t)\geq\kappa_{\max}
{}\land\ \kappa_t(a_t)-\kappa_t(\bar a_t)\geq\epsilon_\kappa
\right],\\
\mathsf V_t
&=\mathbb I\!\left[
\frac{|\Omega_t^{\mathrm{low}}\cap\Omega_t^{\mathrm{vis}}|}
{\max\{1,|\Omega_t^{\mathrm{low}}|\}}\geq\eta
\right].
\end{aligned}
\label{eq:drone_query_criteria}
\end{equation}
Here \(\kappa_{\max}\) bounds acceptable action cost and \(\epsilon_\kappa\) sets the minimum degradation relative to the nominal action. Thus \(\mathsf D_t\) identifies when insufficient reliability hinders navigation, while \(\mathsf V_t\) tests whether new depth is expected to cover enough of the responsible region. Their conjunction gives the observation decision \(q_t\) in Eq.~\eqref{eq:observation_policy}.

\section{Experiments}
\label{sec:experiments}

\subsection{Experimental Setup}
\label{sec:experimental_setup}

\begin{figure}[!t]
\centering
\includegraphics[width=\linewidth]{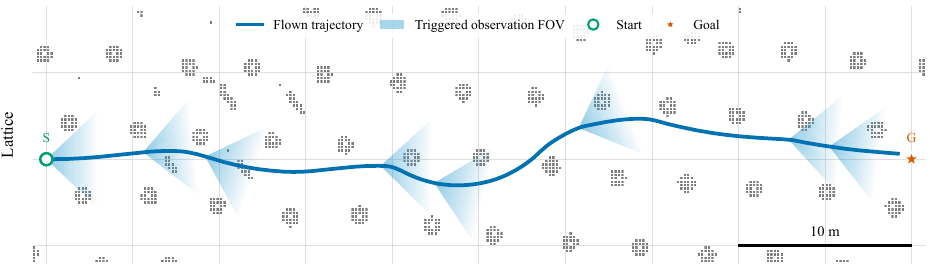}\par\vspace{1pt}
\includegraphics[width=\linewidth]{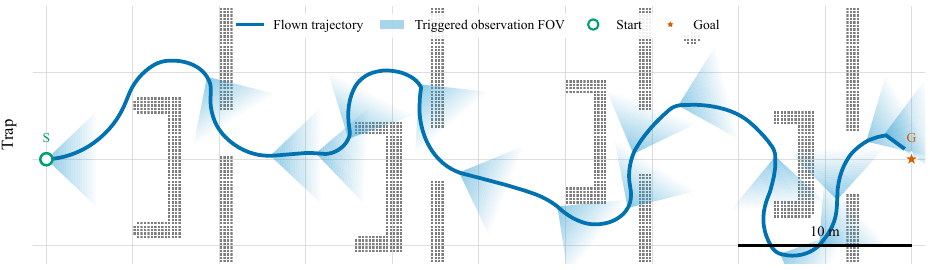}
\caption{\textbf{Representative ALONE rollouts.} Top-down views show executed trajectories from start to goal in Lattice (top) and Trap (bottom); viewing cones mark triggered observations.}
\label{fig:closed_loop_rollouts}
\end{figure}

\paragraph{Scene families}
We use two procedurally generated scenes, illustrated by the representative rollouts in Fig.~\ref{fig:closed_loop_rollouts}. \emph{Lattice} uses a repeating obstacle arrangement with small positional perturbations to test structural continuation and accumulated prediction error between observations. \emph{Trap} hides randomly positioned passages behind occluders to test the representation of unknown structure and its resolution by fusing new evidence.

\paragraph{Training and evaluation protocol}
We generate training and simulated evaluation data in Flightmare, a modular quadrotor simulator~\cite{song2021flightmare}. Expert trajectories provide depth, pose, relative motion, and dense occupancy supervision; all learned methods use identical training data. The ego-centric volume spans \(x,y\in[-10,10]\)\,m and \(z\in[-2,2]\)\,m with \((n_x,n_y,n_z)=(200,200,16)\) voxels. ALONE uses a \((h,w)=(50,50)\) latent map with \(K=512\) codes, and the depth camera has an \(87^\circ\times58^\circ\) field of view. Recurrent training uses \(T=6\) in Eqs.~\eqref{eq:structural_objective} and~\eqref{eq:adequacy_objective}, with \(p_{\mathrm{obs}}=0.2\) in Eq.~\eqref{eq:masked_filter_update}.

For open-loop evaluation, all methods receive the same ten 20-step action--observation sequences per scene, with observations at Steps~0, 4, 8, 12, and~16. We evaluate the posterior at observation steps and the prior otherwise. Closed-loop evaluation uses 100 trials per method in each scene, with a 50\,m start--goal separation. We set \(\tau=0.97\), requiring at least 97\% predicted probability that the spatial estimate is correct at each queried voxel. The navigation policy runs at 10\,Hz. Belief updates are event-driven to reflect structural changes ahead while matching the motion increments used in recurrent training. Without a new observation, ALONE propagates the belief after either 3\,m of world-frame translation or a \(60^\circ\) yaw change since the previous update. A requested observation instead triggers an immediate observation-conditioned update. In addition to Eq.~\eqref{eq:observation_policy}, an observation is forced after at most five observation-free inference steps to remain within the training distribution. Trials terminate upon reaching the goal, collision, 100\,s without progress, or a lateral deviation of 15\,m from the start--goal line.

\begin{table}[!t]
\caption{\textbf{End-to-end closed-loop navigation.}}
\label{tab:closed_loop_navigation}
\centering
\scriptsize
\renewcommand{\arraystretch}{1.05}
\setlength{\tabcolsep}{-0.4pt}
\begin{tabular*}{\columnwidth}{@{\extracolsep{\fill}}llcccc@{}}
\toprule
& & \multicolumn{2}{c}{Navigation performance}
& \multicolumn{2}{c}{Observation demand} \\
\cmidrule(lr){3-4}\cmidrule(l){5-6}
Scene & Method & Succ. rate$\uparrow$ & Duration (s)$\downarrow$ & Obs. count$\downarrow$ & Obs. rate$\downarrow$ \\
\midrule
\multirow[c]{4}{*}{Lattice}
    & \bfseries ALONE & \bfseries\boldmath $98\%$ & \bfseries\boldmath $79.7\,[90.8]$ & \bfseries\boldmath $7\,[12]$ & \bfseries\boldmath $0.009\,[0.014]$ \\
    & OccMap-FullRate & $96\%$ & $99.4\,[139.2]$ & $981\,[1429]$ & $1$ \\
    & OccMap-Periodic-0.5 & $82\%$ & $161.2\,[231.8]$ & $314\,[457]$ & $0.2$ \\
    & OccMap-Periodic-1 & $64\%$ & $227.6\,[305.2]$ & $230\,[311]$ & $0.1$ \\
\midrule
\multirow[c]{4}{*}{Trap}
    & \bfseries ALONE & \bfseries\boldmath $97\%$ & \bfseries\boldmath $121.4\,[160.6]$ & \bfseries\boldmath $16\,[39]$ & \bfseries\boldmath $0.013\,[0.031]$ \\
    & OccMap-FullRate & $\mathbf{97\%}$ & $224.6\,[301.0]$ & $2266\,[3031]$ & $1$ \\
    & OccMap-Periodic-0.5 & $90\%$ & $233.3\,[291.6]$ & $463\,[574]$ & $0.2$ \\
    & OccMap-Periodic-1 & $28\%$ & $290.7\,[344.2]$ & $305\,[344]$ & $0.1$ \\
\bottomrule
\end{tabular*}
\par\vspace{1pt}
\begin{minipage}{\columnwidth}
\scriptsize OccMap periodic suffixes denote observation periods in seconds. Success rate uses all rollouts. Other metrics use successful rollouts only and are reported as median [90th percentile].
\end{minipage}
\end{table}

\begin{figure*}[!t]
\centering
\includegraphics[width=\textwidth]{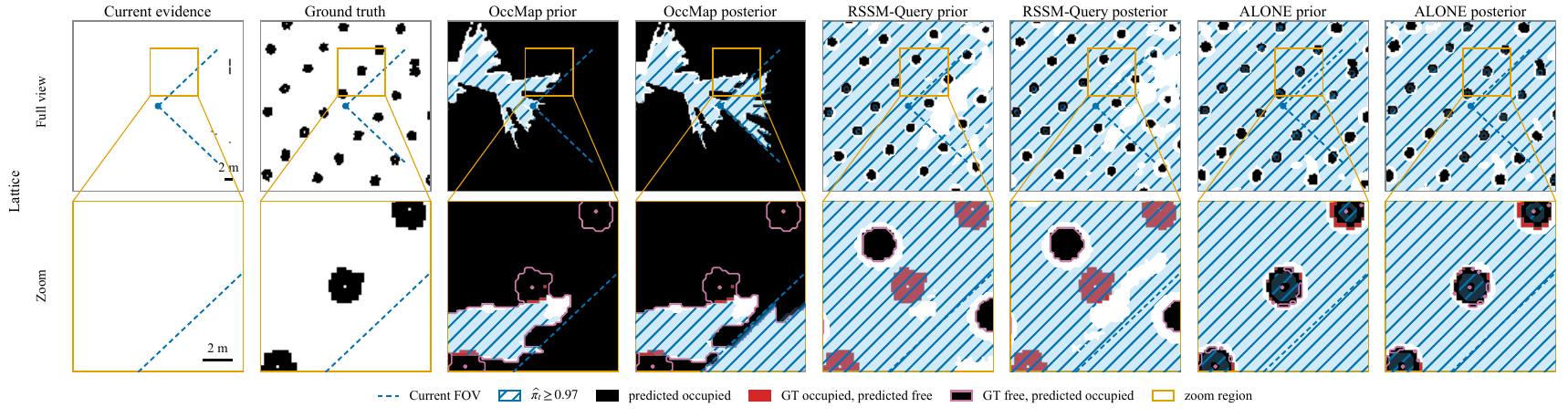}
\caption{\textbf{Qualitative spatial estimates in Lattice.}
Priors precede the new observation; posteriors incorporate it. Dashed blue lines bound the current field of view (FOV); blue hatching marks \(\widehat\pi_t\geq0.97\); black denotes predicted occupancy. Against ground truth (GT), red marks occupied voxels predicted free and magenta marks free voxels predicted occupied. The lower row enlarges the orange regions.}
\label{fig:qualitative_belief}
\end{figure*}

\paragraph{Baselines}
Comparisons comprise three groups. \emph{System-level baselines} (Sec.~\ref{sec:end_to_end_navigation}) model conventional mapping-based navigation as depth-updated probabilistic occupancy mapping with repeated motion planning~\cite{elfes1989occupancy,hornung2013octomap}. \textbf{OccMap-FullRate} and \textbf{OccMap-Periodic} update the map at every decision step and at fixed intervals, respectively, using the same decision rate and motion-planning module as ALONE. \emph{Representation-level baselines} (Sec.~\ref{sec:acting_from_belief_evaluation}) compare ALONE with \textbf{OccMap} and a recurrent state-space model equipped with spatial estimates and query-space reliability maps (\textbf{RSSM-Query}), using shared histories, motion-planning queries, outputs, and supervision. For OccMap, we define reliability as \(\max(p_{\mathrm{occ}},1-p_{\mathrm{occ}})\), which is the model-implied probability that its binary MAP occupancy prediction is correct. RSSM-Query trains the same post-hoc reliability decoder as ALONE. \emph{Observation-policy baselines} (Sec.~\ref{sec:looking_when_needed_evaluation}) keep ALONE's belief and motion-planning module fixed and replace only its observation policy with \textbf{ALONE-Periodic} or \textbf{ALONE-Uncertainty}. ALONE-Uncertainty requests observations based only on belief uncertainty, without evaluating their effect on the selected action.

\paragraph{Metrics}
Closed-loop evaluation measures navigation performance by success rate and flight duration, and observation demand by observation count and observation rate, the fraction of decision steps requiring a new depth image. Open-loop evaluation uses three groups of metrics. Spatial-estimate quality is measured by occupied/free intersection-over-union (IoU) and same-time prior--posterior gain. Reliability calibration is measured by Expected Calibration Error (ECE). Motion-planning relevance is measured by reliable coverage and occupied recall within reliable regions; both are reported in five 2\,m bins over the forward, flight-level query region.

\subsection{End-to-End Navigation under Intermittent Perception}
\label{sec:end_to_end_navigation}

The primary question is whether ALONE reduces navigation observations without sacrificing closed-loop performance relative to OccMap-FullRate and OccMap-Periodic (Sec.~\ref{sec:experimental_setup}). ALONE retains the same decision rate as all baselines while reducing observation frequency.

\paragraph{System-level navigation}
In both scenes, ALONE maintains a success rate comparable to OccMap-FullRate while reducing flight duration. The reduction is especially pronounced in \emph{Trap}, where ALONE represents uncertainty about occluded passages and fuses newly acquired evidence into the maintained belief before committing to a route (Table~\ref{tab:closed_loop_navigation}). ALONE consequently follows smooth trajectories without blindly exploring for passages that remain occluded (Fig.~\ref{fig:closed_loop_rollouts}).

\begin{table}[!t]
\caption{\textbf{Occupancy IoU and reliable-region coverage.}}
\label{tab:main_belief_quality}
\centering
\scriptsize
\renewcommand{\arraystretch}{1.05}
\setlength{\tabcolsep}{3pt}
\begin{tabular*}{\columnwidth}{@{\extracolsep{\fill}}llccccc@{}}
\toprule
& & \multicolumn{2}{c}{All valid voxels} & \multicolumn{3}{c}{$\widehat\pi_t\geq0.97$} \\
\cmidrule(lr){3-4}\cmidrule(l){5-7}
Scene & Method & Occ. IoU$\uparrow$ & Free IoU$\uparrow$ & Coverage$\uparrow$ & Occ. IoU$\uparrow$ & Free IoU$\uparrow$ \\
\midrule
\multirow{3}{*}{Lattice} & ALONE & $\mathbf{0.768}$ & $\mathbf{0.891}$ & $\mathbf{0.632}$ & $\mathbf{0.981}$ & $\mathbf{0.993}$ \\
 & OccMap & $0.332$ & $0.134$ & $0.065$ & $0.043$ & $0.956$ \\
 & RSSM-Query & $0.533$ & $0.774$ & $0.443$ & $0.859$ & $0.959$ \\
\midrule
\multirow{3}{*}{Trap} & ALONE & $\mathbf{0.692}$ & $\mathbf{0.854}$ & $\mathbf{0.484}$ & $\mathbf{0.958}$ & $\mathbf{0.984}$ \\
 & OccMap & $0.332$ & $0.141$ & $0.064$ & $0.032$ & $0.976$ \\
 & RSSM-Query & $0.578$ & $0.794$ & $0.339$ & $0.938$ & $0.979$ \\
\bottomrule
\end{tabular*}
\par\vspace{1pt}
\begin{minipage}{\columnwidth}
\scriptsize The first two IoUs use the complete volume under the common validity mask. Under \(\widehat\pi_t\geq0.97\), coverage is the retained fraction; each class IoU uses the predicted--true union within that subset.
\end{minipage}
\end{table}

\paragraph{Observation demand}
ALONE requests depth at median rates of 0.9\% and 1.3\% of decision steps, reducing observation demand by up to two orders of magnitude while preserving navigation performance. Directly reducing OccMap's observation frequency instead lowers success and increases flight duration (Table~\ref{tab:closed_loop_navigation}). ALONE concentrates observations near obstacles and turns while traversing lattice corridors from the propagated belief (Fig.~\ref{fig:closed_loop_rollouts}), potentially leaving the shared sensor available between requests.

\subsection{Acting from Belief: Spatial Prediction, Correction, and Reliability}
\label{sec:acting_from_belief_evaluation}

Against OccMap and RSSM-Query (Sec.~\ref{sec:experimental_setup}), the analysis tests the spatial estimate, its correction by new evidence, and its predicted reliability.

\paragraph{Spatial-estimate accuracy and reliability calibration}
ALONE preserves the unobserved lattice and infers occluded passage structure from the available history, whereas OccMap leaves them unresolved and RSSM-Query can distort obstacle layout (Fig.~\ref{fig:qualitative_belief}); its spatial estimate also attains the highest occupied/free IoU in both scenes (Table~\ref{tab:main_belief_quality}).

\begin{table}[!t]
\caption{\textbf{Prior--posterior occupancy IoU gain.}}
\label{tab:prior_posterior_quality}
\centering
\scriptsize
\renewcommand{\arraystretch}{1.05}
\begin{tabular*}{\columnwidth}{@{\extracolsep{\fill}}llccc@{}}
\toprule
Scene & Metric & Prior & Posterior & Paired gain \\
\midrule
\multirow{2}{*}{Lattice} & Occ. IoU & $0.770$ & $0.785$ & $\mathbf{0.015}$ \\
                         & Free IoU & $0.892$ & $0.898$ & $\mathbf{0.007}$ \\
\midrule
\multirow{2}{*}{Trap} & Occ. IoU & $0.681$ & $0.701$ & $\mathbf{0.020}$ \\
                      & Free IoU & $0.842$ & $0.853$ & $\mathbf{0.011}$ \\
\bottomrule
\end{tabular*}
\par\vspace{1pt}
\begin{minipage}{\columnwidth}
\scriptsize Only observation steps are included, with prior and posterior evaluated as same-time pairs. Values are mean IoU; paired gain is posterior minus prior.
\end{minipage}
\end{table}

New evidence consistently corrects the predicted belief. At the same observed steps, ALONE's posterior improves occupied and free IoU over its prior in both scenes (Table~\ref{tab:prior_posterior_quality}), matching the correction property in Eq.~(12).

\begin{figure}[!t]
\centering
\includegraphics[width=0.5\linewidth]{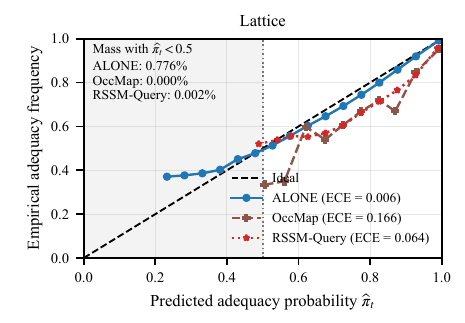}%
\includegraphics[width=0.5\linewidth]{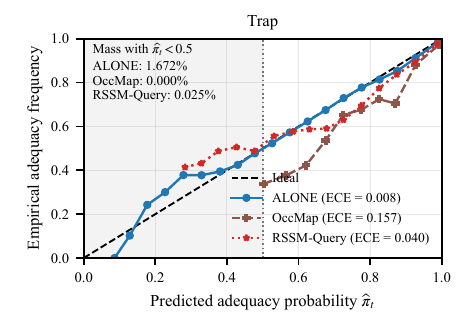}
\caption{\textbf{Query-space reliability calibration.}
Empirical correctness frequency versus predicted correctness probability \(\widehat\pi_t\) in Lattice (left) and Trap (right); the diagonal is ideal, and the inset reports the voxel mass assigned probability below 0.5. Expected Calibration Error (ECE), the bin-weighted absolute calibration gap, is lowest for ALONE.}
\label{fig:adequacy_calibration}
\end{figure}

\begin{figure}[!t]
\centering
\includegraphics[width=0.5\linewidth]{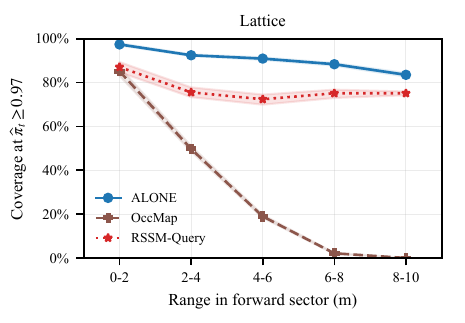}%
\includegraphics[width=0.5\linewidth]{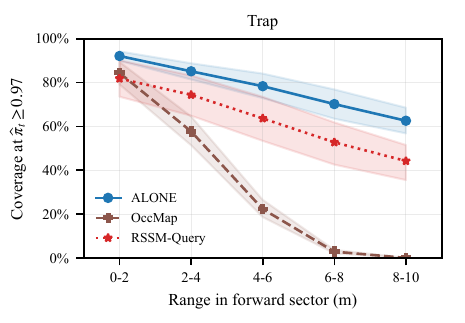}\par\vspace{1pt}
\includegraphics[width=0.5\linewidth]{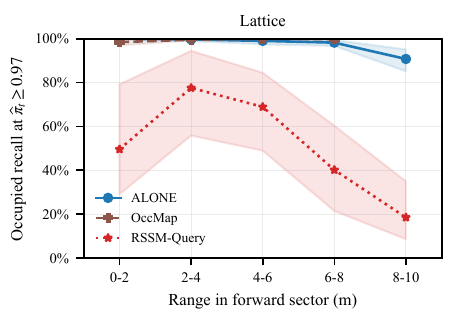}%
\includegraphics[width=0.5\linewidth]{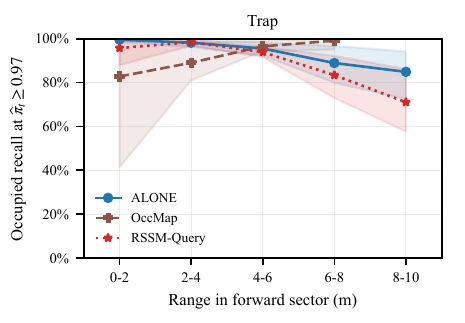}
\caption{\textbf{Reliable occupancy for the motion-planning module.}
Lattice/Trap are left/right. Rows show coverage under \(\widehat\pi_t\geq0.97\) and occupied recall within that subset, binned by forward distance; shading gives 95\% bootstrap confidence intervals over rollouts.}
\label{fig:forward_open_loop_quality}
\end{figure}

ALONE provides a spatially resolved reliability map that indicates where the spatial estimate meets planning requirements. ALONE predicts reliability that agrees with empirical correctness (Fig.~\ref{fig:adequacy_calibration}) and, at \(\widehat{\pi}_t\geq0.97\), achieves the broadest coverage and highest occupied/free IoU (Table~\ref{tab:main_belief_quality}). In the forward region used for collision checking, it also maintains high occupied recall over substantially broader coverage than OccMap, while RSSM-Query misses more obstacles (Fig.~\ref{fig:forward_open_loop_quality}). Reliable coverage declines gradually in \emph{Lattice} as positional errors accumulate, but sharply in \emph{Trap} because hidden passages remain unresolved until observed. The navigation policy can therefore continue selecting actions where the estimate is reliable and request visible evidence when insufficient reliability hinders action selection.

\paragraph{Spatial-inference ablations}

\begin{figure}[!t]
\centering
\includegraphics[width=0.5\linewidth]{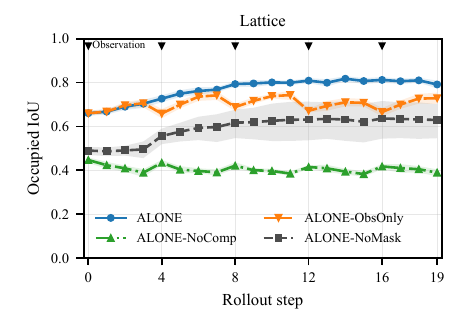}%
\includegraphics[width=0.5\linewidth]{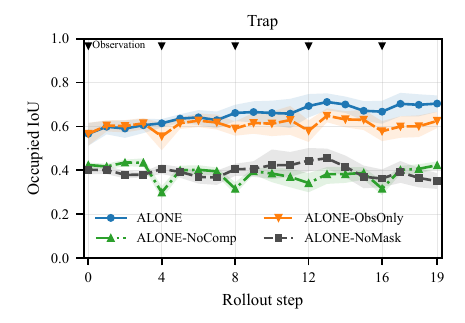}
\caption{\textbf{Spatial-inference ablation over open-loop rollouts.}
Lattice/Trap are left/right. Black markers indicate observation steps; shading gives 95\% bootstrap confidence intervals over rollouts. All variants receive the same action--observation sequence.}
\label{fig:spatial_inference_ablation}
\end{figure}

We ablate learned spatial completion (ALONE-NoComp), history during observation updates (ALONE-ObsOnly), and observation-free training transitions (ALONE-NoMask); each ablation degrades occupied IoU (Fig.~\ref{fig:spatial_inference_ablation}). All three mechanisms thus support accurate spatial prediction and correction under intermittent perception.

\begin{table}[!t]
\caption{\textbf{IoU gains from learned geometric priors.}}
\label{tab:scene_ablation}
\centering
\scriptsize
\renewcommand{\arraystretch}{1.08}
\setlength{\tabcolsep}{4pt}
\sisetup{separate-uncertainty=true,detect-weight=true}
\begin{tabular*}{\columnwidth}{@{\extracolsep{\fill}}ll
S[table-format=1.3]
S[table-format=1.3]
S[table-format=+1.3]@{}}
\toprule
{Scene} & {Metric} & {ALONE} & {ALONE-NoComp} & {Paired gain} \\
\midrule
\multirow{2}{*}{Lattice} & Occ. IoU & 0.766 & 0.405 & \bfseries +0.361 \\
 & Free IoU & 0.890 & 0.722 & \bfseries +0.168 \\
\midrule
\multirow{2}{*}{Random} & Occ. IoU & 0.541 & 0.447 & \bfseries +0.094 \\
 & Free IoU & 0.752 & 0.707 & \bfseries +0.044 \\
\bottomrule
\end{tabular*}
\par\vspace{1pt}
\begin{minipage}{\columnwidth}
\scriptsize Values are mean IoU; paired gain is ALONE minus ALONE-NoComp.
\end{minipage}
\end{table}

We ablate the periodic structure of \emph{Lattice} using an auxiliary \emph{Random} scene that retains the same obstacle geometry but randomizes its arrangement; this ablation markedly reduces completion gains (Table~\ref{tab:scene_ablation}). Learned geometric priors thus exploit predictable structure for spatial completion.

\paragraph{Closed-loop comparison}

\begin{table}[!t]
\caption{\textbf{Closed-loop comparison of spatial representations.}}
\label{tab:closed_loop_model_comparison}
\centering
\scriptsize
\renewcommand{\arraystretch}{1.05}
\setlength{\tabcolsep}{-0.4pt}
\begin{tabular*}{\columnwidth}{@{\extracolsep{\fill}}llcccc@{}}
\toprule
& & \multicolumn{2}{c}{Navigation performance}
& \multicolumn{2}{c}{Observation demand} \\
\cmidrule(lr){3-4}\cmidrule(l){5-6}
Scene & Method & Succ. rate$\uparrow$ & Duration (s)$\downarrow$ & Obs. count$\downarrow$ & Obs. rate$\downarrow$ \\
\midrule
\multirow[c]{3}{*}{Lattice}
    & \bfseries ALONE & $98\%$ & $79.7\,[90.8]$ & $7\,[12]$ & $0.009\,[0.014]$ \\
    & OccMap & $68\%$ & $178.5\,[278.5]$ & $60\,[90]$ & $0.032\,[0.040]$ \\
    & RSSM-Query & $59\%$ & $82.3\,[86.9]$ & $7\,[10]$ & $0.008\,[0.012]$ \\
\midrule
\multirow[c]{3}{*}{Trap}
    & \bfseries ALONE & $97\%$ & $121.4\,[160.6]$ & $16\,[39]$ & $0.013\,[0.031]$ \\
    & OccMap & $19\%$ & $299.6\,[338.0]$ & $121\,[148]$ & $0.043\,[0.055]$ \\
    & RSSM-Query & $75\%$ & $115.2\,[163.0]$ & $18\,[38]$ & $0.016\,[0.023]$ \\
\bottomrule
\end{tabular*}
\par\vspace{1pt}
\begin{minipage}{\columnwidth}
\scriptsize OccMap and RSSM-Query use the same navigation policy as ALONE. Success rate uses all rollouts. Other metrics use successful rollouts only and are reported as median [90th percentile].
\end{minipage}
\end{table}

For closed-loop comparison, OccMap and RSSM-Query use ALONE's navigation policy for action selection and observation decisions. Without environment completion, OccMap requires longer flights and more observations in both scenes, with blind exploration causing frequent timeouts, especially in \emph{Trap}. RSSM-Query achieves observation counts comparable to those of ALONE, but overconfidence in incorrect regions substantially lowers its success rate. Combining environment completion with calibrated reliability, ALONE best balances navigation success, flight duration, and observation demand.

\subsection{Looking When Needed: Observation-Policy Evaluation}
\label{sec:looking_when_needed_evaluation}

We compare ALONE's observation policy with periodic and uncertainty-only alternatives (Sec.~\ref{sec:experimental_setup}). This comparison asks whether ALONE observes at more useful times---when low reliability would alter the selected action and the affected region is expected to be visible to the depth camera.

\begin{table}[!t]
\caption{\textbf{Observation-policy ablation.}}
\label{tab:observation_trigger_ablation}
\centering
\scriptsize
\renewcommand{\arraystretch}{1.05}
\setlength{\tabcolsep}{1.0pt}
\begin{tabular*}{\columnwidth}{@{\extracolsep{\fill}}llcccc@{}}
\toprule
& & \multicolumn{2}{c}{Navigation performance}
& \multicolumn{2}{c}{Observation demand} \\
\cmidrule(lr){3-4}\cmidrule(l){5-6}
Scene & Method & Succ. rate$\uparrow$ & Duration (s)$\downarrow$ & Obs. count$\downarrow$ & Obs. rate$\downarrow$ \\
\midrule
\multirow[c]{6}{*}{Lattice}
 & \textbf{ALONE} & $98\%$ & $79.7\,[90.8]$ & $7\,[12]$ & $0.009\,[0.014]$ \\
 & ALONE-Periodic-2 & $100\%$ & $79.6\,[104.7]$ & $41\,[53]$ & $0.05$ \\
 & ALONE-Periodic-4 & $99\%$ & $78.7\,[99.8]$ & $20\,[25]$ & $0.025$ \\
 & ALONE-Periodic-8 & $97\%$ & $80.5\,[97.0]$ & $11\,[13]$ & $0.013$ \\
 & ALONE-Periodic-16 & $98\%$ & $86.4\,[142.4]$ & $6\,[9]$ & $0.006$ \\
 & ALONE-Uncertainty & $100\%$ & $80.1\,[112.9]$ & $21\,[35]$ & $0.024\,[0.039]$ \\
\midrule
\multirow[c]{6}{*}{Trap}
 & \textbf{ALONE} & $97\%$ & $121.4\,[160.6]$ & $16\,[39]$ & $0.013\,[0.031]$ \\
 & ALONE-Periodic-2 & $100\%$ & $110.2\,[133.2]$ & $56\,[67]$ & $0.05$ \\
 & ALONE-Periodic-4 & $100\%$ & $115.8\,[141.2]$ & $30\,[36]$ & $0.025$ \\
 & ALONE-Periodic-8 & $96\%$ & $131.8\,[191.3]$ & $17\,[25]$ & $0.013$ \\
 & ALONE-Periodic-16 & $53\%$ & $284.1\,[397.5]$ & $18\,[26]$ & $0.006$ \\
 & ALONE-Uncertainty & $97\%$ & $130.6\,[168.2]$ & $32\,[42]$ & $0.024\,[0.032]$ \\
\bottomrule
\end{tabular*}
\par\vspace{1pt}
\begin{minipage}{\columnwidth}
\scriptsize Periodic suffixes denote observation periods in seconds. Success rate uses all rollouts. Other metrics use successful rollouts only and are reported as median [90th percentile].
\end{minipage}
\end{table}

A periodic observation policy imposes a fixed trade-off: frequent observations waste sensing when the selected action is already supported by reliable spatial estimates, whereas infrequent observations delay progress, leave the drone without necessary evidence, and increase the risk of failure, as shown by the sharp degradation of Periodic-16 in \emph{Trap}. Its operating point must be tuned in advance; ALONE instead finds this balance online by adapting observation timing to the needs of current action selection. The uncertainty-only observation policy is more conservative, at least doubling the median observation count in both scenes.

\subsection{Real-World Drone Navigation}
\label{sec:real_world}

We evaluate ALONE with intermittent depth observations on a drone equipped with Intel RealSense D435i and T265 sensors in an indoor \emph{Lattice} environment (Fig.~\ref{fig:exp_overview}). D435i depth corrects the belief, while T265 visual--inertial relative-motion estimates propagate it, matching the simulation inputs. For real-world flight, we use a more conservative configuration than in simulation, requesting observations sooner to limit belief extrapolation and prevent boundary-wall collisions in the confined space. Each trial starts at a random position and requires 10~m of forward navigation. ALONE completes all 10 trials while acquiring depth at only 7.0\% of decision steps, showing that the maintained belief supports action selection and flight between observations with substantially reduced observation demand.

\section{Conclusion and Future Work}
\label{conclusion}

We presented ALONE, a Bayesian spatial world model that uses learned geometric priors to maintain spatial predictions between observations and reliability to guide when perception resumes. In two simulated scene families, ALONE achieves 98\% and 97\% navigation success at a 10 Hz decision rate, with median observation rates of 0.9\% and 1.3\% of decision steps in successful trials, respectively. Real-world flight experiments further demonstrate navigation from the maintained belief between depth observations, with all 10 trials completed successfully. These results support substantially reduced observation demand in the tested environments, potentially freeing the shared sensor for other tasks. Future work will investigate explicit sensor allocation between navigation and other tasks, extend ALONE to dynamic environments, and incorporate semantic knowledge to infer complete object structure from partial observations.

\bibliographystyle{IEEEtran}
\bibliography{references}

\end{document}